# Deep Learning for Automated Quality Assessment of Color Fundus Images in Diabetic Retinopathy Screening

Sajib Kumar Saha, PhD

Australian e-Health Research Centre, CSIRO, Perth, WA 6014, Australia

Basura Fernando, PhD

ACRV, The Australian National University, Canberra ACT 0200, Australia

Jorge Cuadros, OD, PhD

[3]University of California, Berkeley, CA, United States

Di Xiao, PhD

Australian e-Health Research Centre, CSIRO, Perth, WA 6014, Australia

Yogesan Kanagasingam, PhD

Australian e-Health Research Centre, CSIRO, Perth, WA 6014, Australia

# Deep Learning for Automated Quality Assessment of Color Fundus Images in Diabetic Retinopathy Screening

**Abstract [192 words]**

Purpose:

To develop a computer based method for the automated assessment of image quality in the context of diabetic retinopathy (DR) to guide the photographer.

Methods:

A deep learning framework was trained to grade the images automatically. A large representative set of 7000 color fundus images were used for the experiment which were obtained from the EyePACS (http://www.eyepacs.com/) that were made available by the California Healthcare Foundation. Three retinal image analysis experts were employed to categorize these images into 'accept' and 'reject' classes based on the precise definition of image quality in the context of DR. A deep learning framework was trained using 3428 images.

Results:

A total of 3572 images were used for the evaluation of the proposed method. The method shows an accuracy of 100% to successfully categorise 'accept' and 'reject' images.

Conclusion:

Image quality is an essential prerequisite for the grading of DR. In this paper we have proposed a deep learning based automated image quality assessment method in the context of DR. The method can be easily incorporated with the fundus image capturing system and thus can guide the photographer whether a recapture is necessary or not.

**[3868 words]**

**1. Introduction**

Color fundus imaging, a non-invasive examination of the eye, is considered as an efficient modality to screen for and diagnose several retina related conditions [1, 2, 3] such as diabetic retinopathy [4], age related macular degeneration [5] etc. Retinal images obtained in a screening program are acquired at different sites, using different cameras that are operated by qualified people who have varying levels of experience [6]. These results in a large variation in image quality and a relatively high percentage of images with inadequate quality for diagnosis. In [7], Abramoff et al. reported that about 12% of the images collected from a screening program are later marked as unreadable by ophthalmologist. Unreadable images require a recapture for diagnosis. However, in many cases like in reading centers in Germany, Australia and USA [7], the image acquisition is time and location independent from its medical assessment. A reacquisition of the images will be time consuming and expensive. Thus assuring sufficient image quality during acquisition is undoubtedly important.

The development of a precise image quality index is not a straightforward task, mainly because quality is a subjective concept which varies even between experts, especially for images that are in the middle of the quality scale. In addition, image quality is dependent upon the type of diagnosis being made. For example, an image with dark regions can be considered as good for glaucoma grading, however, can be considered bad for diabetic retinopathy grading [8].

The proposed image quality metric relies on the definition of acceptable image quality proposed in [9] for diabetic retinopathy (DR); which eventually sets DR as the applicable area of the proposed method. According to [9], image quality can be categorized as 'good', 'adequate' and 'inadequate'.

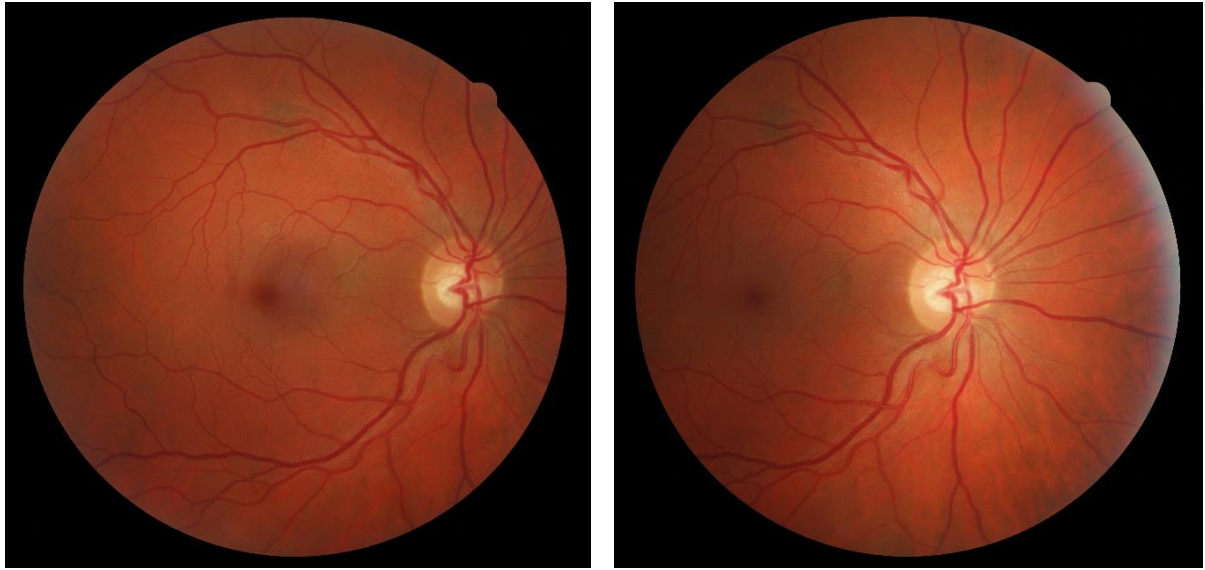

(a) (b)

Figure 1: A typical (a) macula centered and (b) optic disk centered images of the same eye.

Table 1: Image quality specification according to UK national screening committee [9].

|  | **Good quality** | **Adequate quality** | **Inadequate quality** |
|---|---|---|---|
| **Macula centered images** | Center of fovea ≤ 1 DD[1] from center of image & vessels clearly visible | Center of fovea > 2 DD from edge of image & vessels | Failure to meet adequate image quality standards. |

---

[1] Disk diameter

| | | | |
|---|---|---|---|
| | within 1DD of center of fovea & vessels visible across > 90% of image | visible within 1DD of center of fovea | |
| **Optic disk (OD) centered images** | Center of disc ≤ 1 DD from center of image & fine vessels clearly visible on surface of disc & vessels visible across > 90% of image | Complete optic disc > 2 DD from edge of image & fine vessels visible on surface of disc | Failure to meet adequate image quality standards. |

In this work we categories image quality into 'accept' and 'reject' class, where 'accept' class include images with 'good' and 'adequate' quality standards; 'reject' class include images of 'inadequate' quality. The reason for choosing only two categories for image quality has been partially biased by the comments of Giancardo et al. in [8] - "human experts are likely to disagree if many categories of image quality are used".

A deep learning framework is trained based on 'accept' and 'reject' image classes. Once the training is done, the framework is applied to categorise test images. It is worth mentioning, even though we had precise definition for image categories, we realised discrepancy between individual subjective opinion for categorising images, specifically for images that are on the border line. This is consistent with the comments of Paulus et al. in [10] "it is an individual decision at which point the image quality becomes too bad for a stable diagnosis". We identified discrepancy between subjects on about 2% of our images and we called these images as 'ambiguous'. The 'ambiguous' image set is not used during training; however, used during testing.

## 1.1. Review of the Retinal Image Quality Assessment Methods

Several approaches have been developed to automatically determine the quality of the retinal images. These approaches can be divided into two groups based on which image parameters/criteria they consider to classify the image quality [11]. The first group is based on generic image quality parameters such as sharpness and contrast. In 1999, Lee et al. [12] proposed an automated retinal image quality assessment method based on global image intensity histogram. In 2001, Lalonde et al. [13] proposed a method based on analysis of the global edge histogram in combination with localized image intensity histograms. Davis et al. [14] proposed a method of quality assessment based on contrast and luminance features. A method based on sharpness and illumination parameters was proposed by Bartling [15] in 2009. Illumination was measured through evaluation of contrast and brightness and the degree of sharpness was calculated from the spatial frequencies of the image. In 2014, Dias et al. [16] proposed a method based on fusion of generic image quality indicators such as image color, focus, contrast, and illumination.

The second group is based on structural information of the image. To the best of our knowledge the first quality assessment method based on eye structure criteria was proposed by Usher et al. [17] in 2003.This method is based on the clarity and area of the detected eye vasculature and achieved a sensitivity of 84.3% and a specificity of 95.0% in a dataset of 1746 retinal images (17, 18). In 2005, Lowell et al. [20] and in 2006, Fleming et al. [18] presented methods that are very specific to retinal image analysis. Both methods classify images by assigning them to a number of quality classes. An analysis was made on the vasculature in a circular area around the macula and the presence of small vessels in this region was used as the indicator of image quality. In [6], Niemeijer et al. proposed a method based on clustering the filter bank response vectors in order to obtain a compact representation of the image structures. The authors tested the method on 2000 images and reported an area of 99.68% under the receiver operating characteristic (ROC) curve. Inspired by the work of Niemeijer et al., in [21], Giancardo et al. proposed a method focused on eye vasculature only. The proposed method achieved an accuracy of 100% on the identification of 'good' images, 83% on 'fair' images, 0% on 'poor' images and 11% on 'outlier' images in a dataset of 84 retinal images. In 2011, Hunter et al. [19] proposed a method based on clarity of retinal vessels within the macula region and contrast between fovea region and retina backgrounds. The method achieved a sensitivity of 100% and a specificity of 93% in a dataset of 200 retinal images.

The two groups of image features mentioned above were first combined in a work by Paulus et al. [10] in 2010. Image structure clustering [6], Heralick features, and sharpness measures based on image gradient magnitudes were used to classify poor quality retinal images. The method achieved a sensitivity of 96.9% and a specificity of 80.0% on a dataset composed of 301 retinal images.

## 1.2. Deep Learning

Deep learning, also known as deep structured learning, hierarchical learning or deep machine learning, is a branch of machine learning based on a set of algorithms that attempt to model high-level abstractions in data by using multiple processing layers [21, 22]. While traditional machine learning approaches rely on hand-crafted features to extract useful information from the image, deep learning aims to employ machine to learn the features by itself [24].

In recent years, deep learning architectures, such as deep convolutional neural networks (CNNs), have gained significant attention in computer vision [25, 26, 27]. Typically, when used for image recognition, deep CNN network architectures take an image as input and feed forward through several convolutional layers. Each convolutional layer consists of several convolutional filters. Let us say there are $N$ number of filters that are trainable from training data. Each convolutional filter is a kernel consisting of several trainable weights. Number of weights depends on the size of the filter. Usually, filters are square shaped, eg. (3×3 or 11×11). Each convolutional layer at level $L$ takes an image of dimensionality $dL$ ($d1 = 3$ at the input layer for RGB) and apply $N$ number of filters to produce $N$ number of feature maps. This convolution operation produces an $N$ dimensional image, one dimension per filter. This $N$ dimensional image is then taken as input to the next convolutional layer at level $d(L+1)$. This process continues for several convolutional layers. Spatial sub-sampling is also applied using spatial max pooling operation, in order to reduce the computation in classification and avoid over fitting. In addition to that, lateral inhibition is employed to avoid training dominant filters. Finally, several fully connected neural network layers are added on top of convolutional layers. Typically, the final layer is for classification, in which softmax function is often implemented. An illustration of Alexnet neural network is shown in Figure 2.

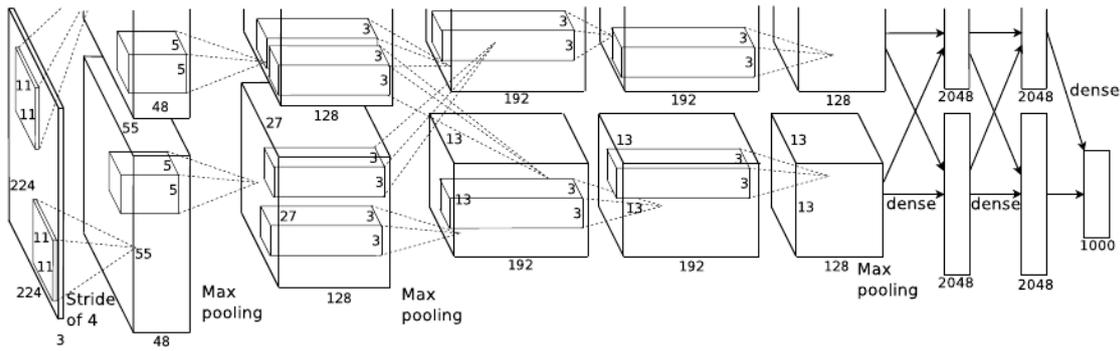

Figure 2: AlexNet convolutional neural network [25].

## 2. Methods

We train a convolution neural network (CNN) based on colour fundus image set that were classified as 'accept' or 'reject' in the context of DR. Once learned, the network is used for two-class image classification task of accepting or rejecting test images.

### 2.1. Materials

A total of 7000 images were obtained from the EyePACS (http://www.eyepacs.com/) which were made available by the California Healthcare Foundation. The images were chosen to cover the greater demographic diversity and different fundus cameras that were found in EyePACS. The images were optic disc (OD) or macula centered.

Three retinal image analysis experts including one ophthalmologist were asked to grade those image as 'accept' and 'reject' based on the definition given in Table 1. A computer platform was developed where each of the experts was shown the images and was asked to provide his/her opinion to 'accept' or 'reject' the images. Even with the precise definition of 'accept' and 'reject' classes, we observed discrepancy between individual subjective opinion especially for images that lied in the border line. The images for which we observed discrepancy between individual subjects were categorized as 'ambiguous'. We identified 147 images (about 2%) as 'ambiguous'; 249 images (about 4%) were classified as 'reject' and rest were classified as 'accept'. The 'accept' and 'reject' classes were used to train the CNN. 50% of the images were used for training and 50% were used for testing. Figure 3 and Figure 4 show exemplary images of the 'accept' and 'reject' classes respectively. Figure 5 shows sample images of the 'ambiguous' class.

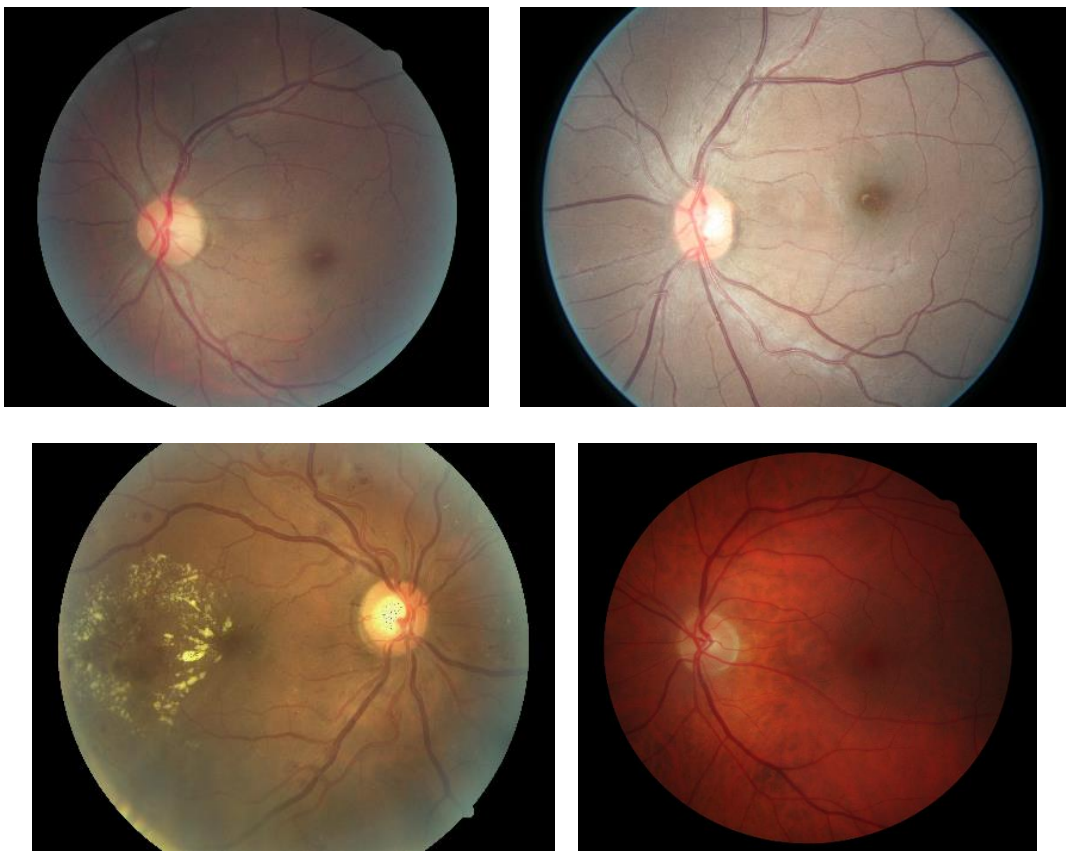

Figure 3: Sample images of the 'accept' class.

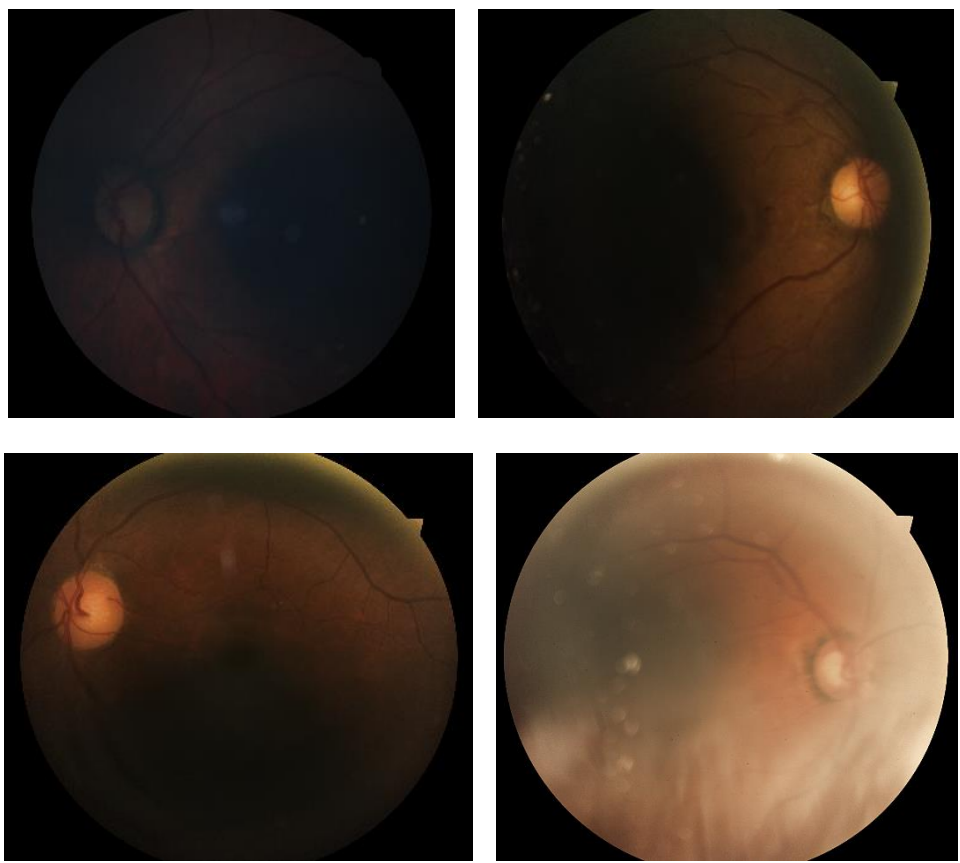

Figure 4: Sample images of the 'reject' class.

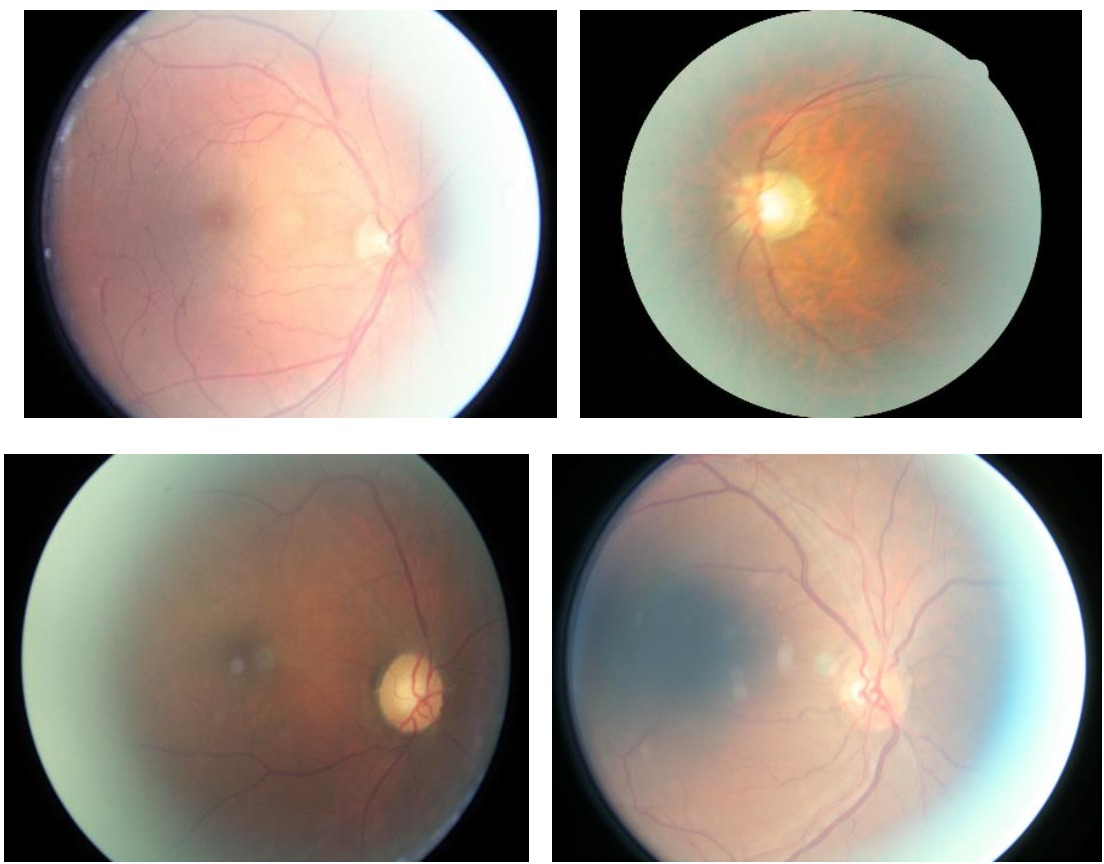

Figure 5: Sample images of the 'ambiguous' class.

## 2.2. Training the Deep Learning Network

We used Alexnet CNN architecture [25]. The network was modified for two-class classification task and we used the hinge loss as the loss function [28]. Our CNN architecture consists of five convolution layers, two fully connected layers and a binary classification layer. Each convolution layer performs the convolution operation on the input images of variable number of channels. The first convolution layer consists of 96 filters of size 11 by 11. The second convolution layer of the CNN consists of 256 filters. The third and fourth convolution layers consist of 384 filters each. The final convolution layer consist of 256 filters. The activation size of the first and the final fully connected layer has 4096 filters in each. The final fully connected layer produces an output of dimensionality 4096 by 1. So the image information is encoded to a 4096 dimensional feature vector and finally the binary classifier makes predictions based on these 4096 dimensional encoding of image data. The input to our network is a 256 by 256 RGB image consisting of 3 channels. The EyePACS images were first cropped so that the dark background surrounding the fundus is little as possible. Following that the images were resized to 256 by 256.

Mathematically, given pairs of training data denoted by $D = \{(x_1,y_1), (x_2,y_2), \ldots (x_n,y_n)\}$ which consist of $n$ training samples, each $x_i$ is an image and $y_i$ is the class label. The final CNN classifier consist of 4096 parameters which we denote by $w$. During the training, we seek the CNN classifier and filter parameters $\theta$ by minimising the hinge loss over the training data as given by the following equation.

$$Loss(\theta, w, D) = \sum_{i=1}^{n} max(0, y_i \times w^T CNN_\theta(x_i)).$$

During the training of CNN, we minimise the hinge loss for binary classification case. Here the CNN function takes each image $x_i$ and returns a vector. Then we take the dot product between this vector and the classifier w to get the classification score. The label for each image is either +1 or -1. If the classifier score is positive and the label is also positive (+1), then there is no loss occurred. However, if the CNN classifier score is negative and the sample ground truth label is positive, then we obtain positive loss and the objective of CNN training is to reduce such losses. To make sure we find good image representations and the classifier, we minimise this hinge loss over the entire training set using stochastic gradient descent.

To obtain probabilistic values for the predictions we need to calibrate the scores. For this task, one can use generalised logistic function. However, in our method we use the scores return by the CNN to make the final predictions.

We used the Caffe reference model [29] to initialise our networks and then fine-tuned the parameters using colour fundus image data. Caffe reference model is the most widely used neural network. These network models were trained on very large image collections such as ImageNet [30]. For fine-tuning we used stochastic gradient decent method and used a variable learning rate starting from 0.01 and ending at 0.0001. We fine-tuned the entire network for 20 epochs.

## 3. Results

The CNN was trained using 3428 color fundus images, the rest 3425 images were used for evaluation. The test dataset had 3302 'accept' images, 123 'reject' images categorised based on subjective evaluation. We computed the accuracy (i.e. the number of correctly classified images divided by the total number of images) and plotted the ROC curve to analyse the performance of the two-class classification task. ROC curves plot the true positive fraction against the false positive fraction. The VLFeat software package [31] was used to generate the ROC curves. Figure 6 shows the ROC plot for the 'accept' class and Figure 7 the ROC plot for the 'reject' class.

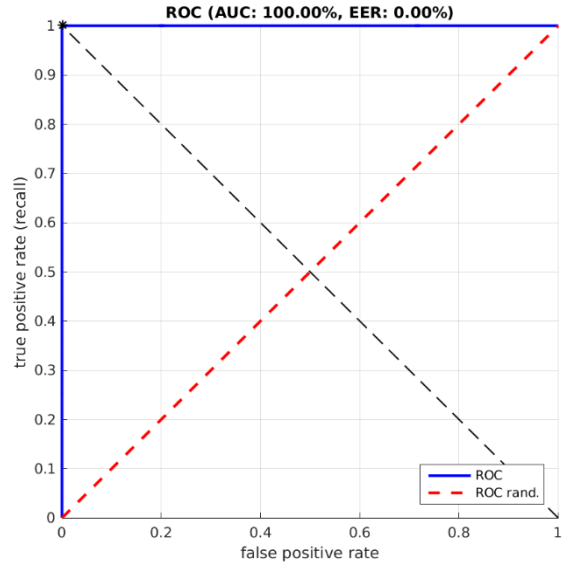

Figure 6: ROC plot of the 'accept' class.

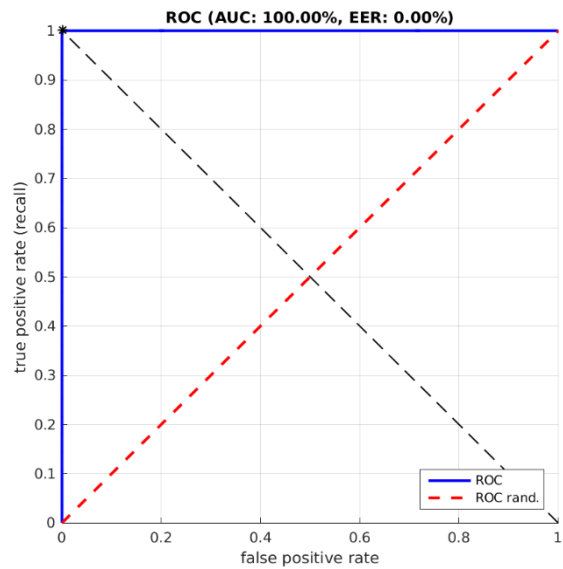

Figure 7: ROC plot of the 'reject' class.

The area under the ROC curve and the accuracy was 100% both for the 'accept' and 'reject' classes.

While we did not use the 'ambiguous' images to train our CNN, we did employ the trained CNN to compute the classification scores for those images. For the 'ambiguous' images, the computed score is far apart from the 'accept' image scores and somewhere in between 'accept' and 'reject' scores. Which eventually means along with 'accept' and 'reject' images, the network is able to detect images that are on the border line. Figure 8 shows the mean classification scores returned by the CNN for the 3 different image categories. The 'accept' images get a score between 0.5 ~ 1.0, the 'reject' images have a score between -2.6 ~ -2.3 and the 'ambiguous' images get a score between -2.1 ~ -1.5.

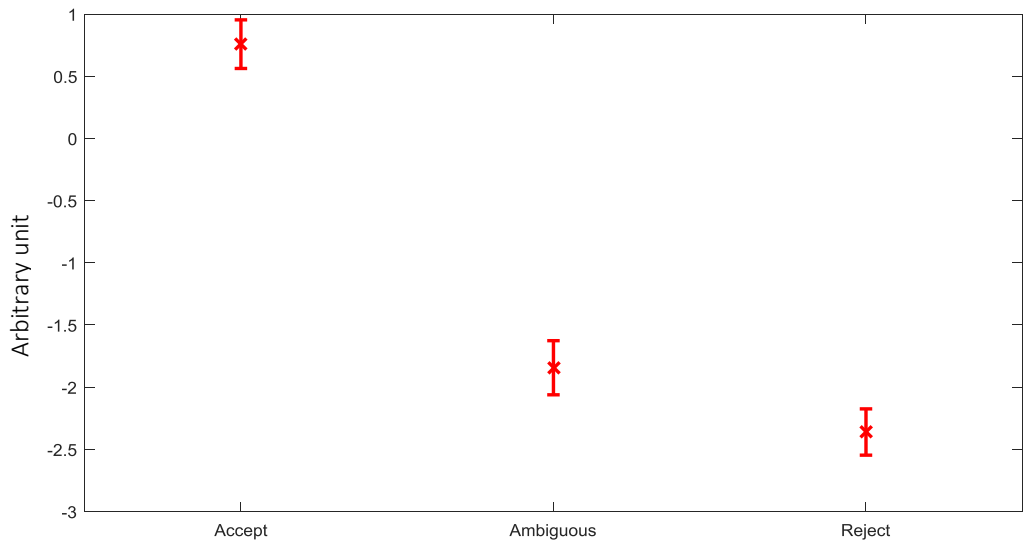

Figure 8: Classification scores returned by the CNN for the 3 different image categories. Whiskers show the standard deviations.

## 4. Discussions and Conclusion

We have proposed a deep learning framework to automatically determine the image quality with the aim to help/assist the photographer to decide on whether a recapture is necessary or not. We have observed that even with precise definition of 'accept' and 'reject' categories, individual subjective opinion varies specially for images that lied in the border line. We identified about 2% of the images where individual subjective opinion varied and we categorised those images as 'ambiguous'. With that finding, we considered two different cases to train the CNN. In one case along with 'accept' and 'reject' classes we also used 'ambiguous' image class to train the network. In the other case we used 'accept', 'reject' classes only to train the network. Our experimental findings have revealed that when 'ambiguous' images are also used for training, it confuses the whole network and overall performance decreases. Figure 9 shows the ROC plots for the 'accept', 'reject' and 'ambiguous' classes. In comparison, when the CNN was trained for the 'accept', 'reject' classes, the accuracy was significantly high (Figures 6 and 7). Hence the proposed method considered training using the 'accept' and 'reject' classes only. It is worth mentioning, the proposed framework returns a score and the scores for the 'ambiguous' images have different range in comparison to that of the 'accept' and 'reject' images. Which eventually means along with accurately identifying 'accept' and 'reject' images, the proposed framework is able to detect images that are on the border line and help/assist the photographer to decide on whether a recapture is necessary or not.

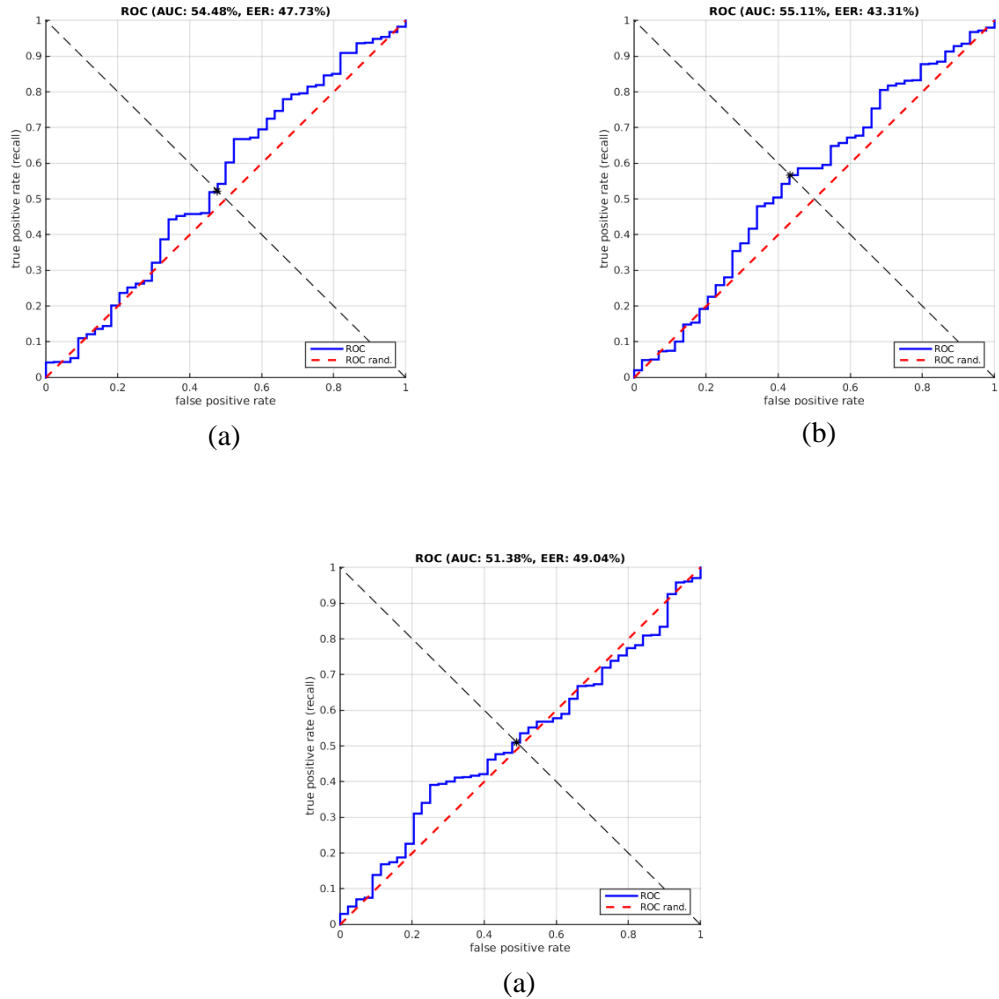

Figure 9: ROC plots of the (a) 'accept', (b) 'reject' and (c) 'ambiguous' class.

The automated image quality verification system proposed in this paper shows excellent results. Two different training setups for the CNN were tested and the best one has been identified. The experiments were conducted on a large, representative set of screening images. The total running time to categorize a given image is about 15 seconds[2]. The software has not been optimized extensively and therefore further increases in speed can be expected.

---

[2] Processor: Intel Core i7 2.90 GHz, RAM: 32 GB

To summarize, image quality assessment is an important problem in large scale DR screening [Niemeijer 2006]. On that perspective, the proposed automated method for retinal image quality assessment provides an efficient way to solve the problem and can easily be incorporated with the DR screening system. The proposed method does not require any previous segmentation of the images in contrast with some other methods and ensures an accuracy of 100% to categorize the 'accept' and 'reject' images.